\pdfoutput=1

\documentclass[11pt]{article}

\usepackage[]{acl}

\usepackage{times}
\usepackage{latexsym}
\usepackage{multirow}
\usepackage{graphicx}

\usepackage[T1]{fontenc}

\usepackage[utf8]{inputenc}

\usepackage{microtype}

\title{Evaluating Contextual Embeddings and their Extraction Layers for Depression Assessment}

\author{Matthew Matero \hspace{1em} Albert Hung \hspace{1em} H. Andrew Schwartz \\
Department of Computer Science\\
Stony Brook University \\
\texttt{\{mmatero, has\}@cs.stonybrook.edu}}

\begin{document}
\maketitle
\begin{abstract}
Recent works have demonstrated ability to assess aspects of mental health from personal discourse. 
At the same time, pre-trained contextual word embedding models have grown to dominate much of NLP but little is known empirically on how to best apply them for mental health assessment. 
Using degree of depression as a case study, we do an empirical analysis on which off-the-shelf language model, individual layers, and combinations of layers seem most promising when applied to human-level NLP tasks. Notably, we find RoBERTa most effective and, despite the standard in past work suggesting the second-to-last or concatenation of the last 4 layers, we find layer 19 (sixth-to last) is at least as good as layer 23 when using 1 layer. Further, when using multiple layers, distributing them across the second half (i.e. Layers 12+), rather than last 4, of the 24 layers yielded the most accurate results.
\end{abstract}

\section{Introduction}

Over the past decade natural language processing (NLP) has increasingly set its sights on interdisciplinary tasks, notably those within the computational social sciences~\cite{sap2014developing,preoctiuc2016modelling,zamani2018residualized}. As more and more language has been generated on social media sites such as Facebook, Twitter, and Reddit, researchers have had a wealth of personal discourse available to them that spans across thousands of users.

Many researchers focus on applying these social media datasets to predict user demographics, personality, or mental health~\cite{matero2019suicide, iyyer2014political, lynn2020hierarchical}. Those predicting facets of mental health, such as depression and suicide risk, can help an over-burdened mental health industry by using automated screening~\cite{coppersmith2018natural}. Often these automated tools can be applied to forums where a user is an active member and their account could be flagged to be brought to the attention of a moderator. Thus, a personalized and potentially early intervention could be provided to the user in question. 

Here, we investigate one prominent aspect of mental health: \textit{degree of depression} (DDep) as measured by answers to an online questionnaire administered to Facebook users. Depression assessment of social media users is of interest for the following reasons: (1) Depression is often highly correlated with suicidal tendencies~\cite{leonard1974depression} with deaths by suicide on the rise~\cite{curtin2016increase} and (2) Automated assessment of depression is of high importance as it is often an under-diagnosed ailment, where such predictions could be useful to screen individuals who are at risk~\cite{eichstaedt2018facebook}. 

While many recent NLP pipelines have moved onto leveraging large pre-trained language models based on the transformer architecture~\cite{vaswani2017attention}, applying these models to human-level analysis, such as predicting a person's states or traits, has received little attention. Even the use of extracted embeddings, often called contextual embeddings, has yet to be fully explored in this level of analysis~\cite{v-ganesan-etal-2021-empirical}. We expand this area of research by investigating how best to leverage the individual layers of off-the-shelf transformer models for depression assessment. Notably, we are interested in going beyond just a single layer and propose a greedy algorithm for selecting layers to extract contextual embeddings and aggregate them for large user-level embeddings.

\textbf{Our contributions include: } (1) A predictive model for depression assessment that out-performs the current state-of-the-art, (2) Evaluation of standard extraction techniques on contextual embeddings and their ability to detect depression levels and (3) Analysis on the effectiveness of layer selection to generate large contextual embedding representations of users.


\section{Related Works}

One of the downsides when modeling mental health data is often that it is very small, with only a few hundred participants per study~\cite{guntuku2017detecting}. However, it is sometimes possible to get around this by using data from Social Media websites where participants can choose to opt in to share past language data and take a small survey or questionnaire~\cite{coppersmith2014measuring}. 
~\citet{schwartz2014towards} applies this technique to Facebook users and evaluates their DDep over a continuous scale (1-5) rather than bucketing users into classes such as mild/moderate/severe. 

Even somewhat recent human-level models in NLP have used bag-of-words style approaches for prediction~\cite{lynn2019tweet, andy2021predicting}, while other areas such as word or document-level tasks have adopted contextual embedding representations~\cite{bao-qiao-2019-transfer, babanejad2020affective, matero-etal-2021-melt-message}. As these are often output from very large models, with hundreds of millions or more parameters, they are able to encode syntactic and semantic information that transfer to downstream tasks either through word or sentence embeddings~\cite{guu2020realm}. 

While there has been some work applying contextual embeddings and transformer language models to human-level predictions, the most in depth has been ~\citet{v-ganesan-etal-2021-empirical} who investigated the use of contextual embeddings in low-data scenarios across various areas including mental health, demographics, and personality assessment. However, they only focus on using the base-size variants with an emphasis on dimensionality reduction techniques to apply contextual embeddings to small datasets (N <= 1000). Here, we work with a medium size dataset of 3 million Facebook posts across 25 thousand users and apply both base and large sized language models, as well as investigate layer selection beyond using just the second to last layer of the model.  

\section{Methods}

\label{sec:methods}

\paragraph*{Task: }
A person's degree of depression score is estimated by their response to a subset of neuroticism questions on a personality assessment through Facebook's MyPersonality app~\cite{schwartz2013personality}. The responses were on a scale of 1 to 5 and averaged together to represent a person's overall degree of depression. Here, we formulate the task of depression assessment as building a single user representation where each status is processed through a language model as a sentence and then all words from a user are avg-pooled. We evaluate our models using mean squared error(MSE) and disattenuated pearson r($r_{dis}$) to account for questionnaire reliability~\cite{lynn-etal-2018-clpsych}. We perform all experiments using the DLATK~\cite{schwartz2017dlatk} library.


\paragraph*{Transformer Language Models: }

From the wide selection of general purpose language models, we select the following: XLNet, RoBERTa, ALBERT and BERT~\cite{yang2019xlnet,liu2019roberta, lan2019albert, devlin2019bert}. These models are chosen as they cover common language model types (e.g. autoregressor vs autoencoder), have been pre-trained on various corpus sizes, and in the case of ALBERT offer a more lightweight footprint in terms of total model parameters. 

When comparing which language model to perform our layer analysis on, we first evaluate performance using only the second to last layer on our held-out test set. This allows us to deduce which model may lead to better application to aggregate human-level predictions. 


\begin{table}[tb]
\centering
\begin{tabular}{lcc}
\hline
    \textbf{Model} & \textbf{$r_{dis}$} & \textbf{MSE}\\ \hline
    \textit{Baselines}&&  \\ 
    \hspace{14pt}Open-Ridge & .507 & .7696 \\
    \hspace{14pt}\citeauthor{schwartz2014towards} & .526 & N/A \\ \hline
        AvgPool-XLNet & .499 & .7728 \\
        AvgPool-BERT &  .528 & .7575 \\ 
        AvgPool-ALBERT & .508 & .7675 \\
        AvgPool-RoBERTa & \textbf{.542*} & \textbf{.7497*}\\
    \hline
    \end{tabular}
    \caption{Performance of extracting embeddings from second to last layer (11) from \textit{base} sized variants of each language model on the held-out test set. Each model is used to encode a 768 dimensional vector for all words that are then averaged to a user representation. \textbf{Bold} indicates best in column and * indicates statistical significance $p < .05$ w.r.t AvgPool-BERT via paired t-test. }
\label{tab:baseline_base}
\end{table}

\begin{table}[tb]
\centering
\begin{tabular}{lccc}
\hline
    \textbf{Model} & \textbf{Hid. Size} &\textbf{$r_{dis}$} & \textbf{MSE}\\ \hline
        RoBERTa-B L11 & 768 & .542 & .7497 \\
        RoBERTa-L L23 & 1024 & \textbf{.543} & \textbf{.7476}\\
        DistilRoBERTa L5 & 768 & .533 & .7545 \\
    \hline
    \end{tabular}
    \caption{Performance of extracting embeddings from second to last layer of RoBERTa variants, which was found to be the best performing among base models, on the held-out test set. DistilRoBERTa is also considered as a small sized alternative. \textbf{Bold} indicates best in column.}
\label{tab:roberta_compare}
\end{table}

\paragraph{Layer Selection: } 
To decide on which layers to extract for our final model, we perform a 10-fold cross-fold validation, for each individual layer or combination of layers. First we select the best performing layer, once found, we then concatenate all other layers to find the best 2-layer combination. This process is iterated on until we reach a number of layers where we cease to see a performance increase via the cross-folds. Once the best performing layers are found via cross-folds, we extract a final test set representation and run the final selection on our held-out test set. When comparing within cross-folds we only compare the MSE, rather than correlation, as that is the metric being optimized as well as being a less noisy evaluation of each model.  

As well as our best performing layer combinations, for a final comparison on the test set, we also evaluate performance of standard layer extraction techniques. This includes the second-to-last layer and the concatenation of the top-4 layers enabling us to validate that our layer selection method and suggested layers are worthwhile. 



\begin{figure}[tb]
    \centering
    \includegraphics[width=1.10\linewidth]{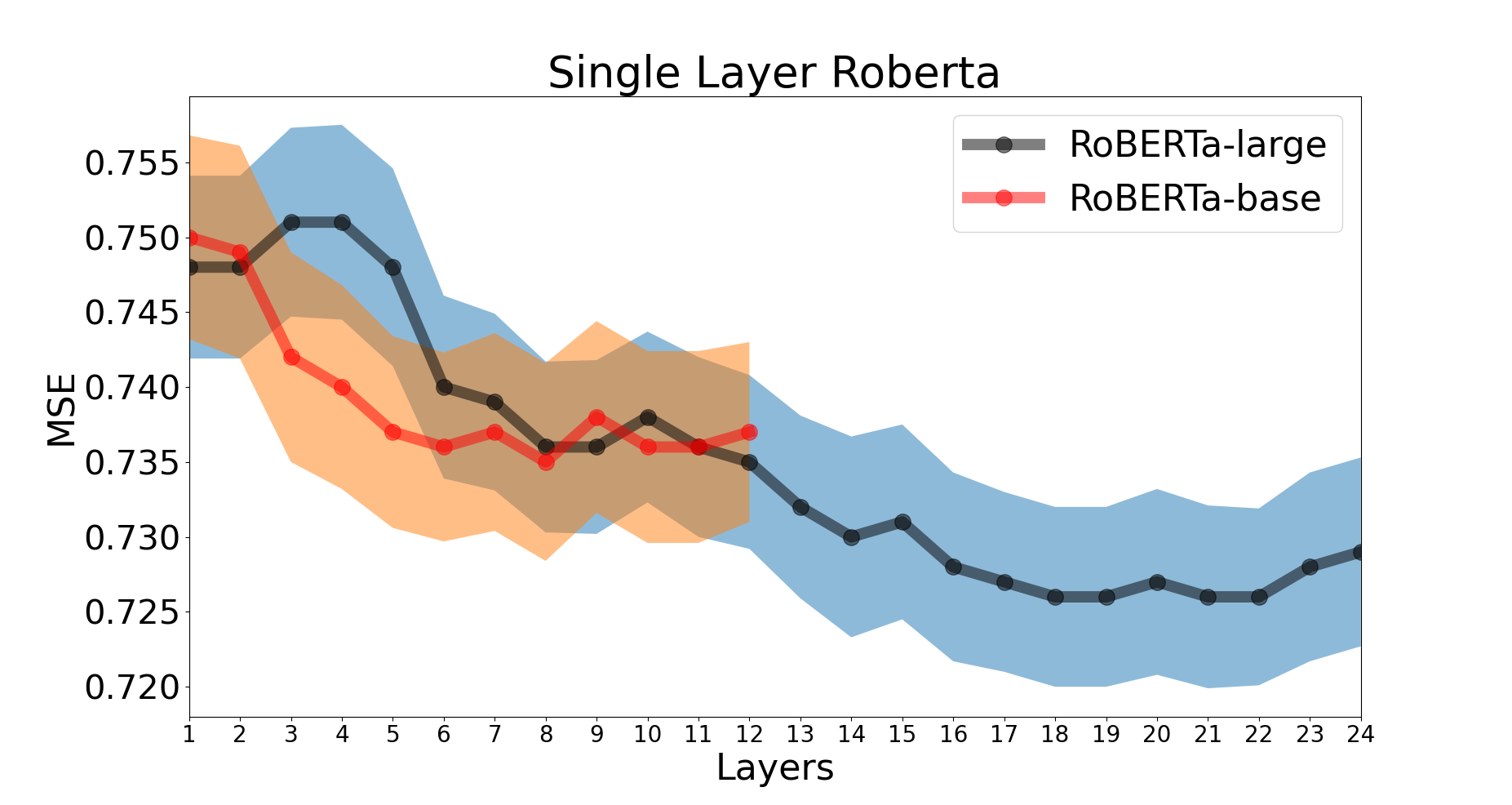}
    \caption{Layer-wise mean squared error performance across the 10-fold validation set with standard error shown by the shaded region for both RoBERTa-base and large. At lower layers (3-6), RoBERTa-base shows a much lower error rate. However at layer 13 and higher of RoBERTa-large there is lower error beyond any available base layer.}
    \label{fig:roberta_layers}
\end{figure}

\begin{table*}[tb]
\begin{center}
\begin{small}
\begin{tabular}{|c|ll|ll|ll|ll|ll|ll|}
\hline
\multirow{1}{*}{\bf{Rank}} &
\multicolumn{2}{c|}{\it{1 Layer}} &
\multicolumn{2}{|c|}{\it{2 Layers}} &
\multicolumn{2}{|c|}{\it{3 Layers}} &
\multicolumn{2}{|c|}{\it{4 Layers}} &
\multicolumn{2}{|c|}{\it{5 Layers}} &
\multicolumn{2}{|c|}{\it{6 Layers}}\\ 
\hline
1 & \textbf{19} & \textbf{0.7257} & \textbf{16} & \textbf{0.7234} & \textbf{24} & \textbf{0.7215} & \textbf{22} & \textbf{0.7208} & \textbf{18} & \textbf{0.7206}  & \textbf{14} & \textbf{0.7207} \\
2 & 18 & 0.7264 & 15 & 0.7241$\downarrow$ & 22 & 0.7216 & 21 & 0.7210 & 17 & 0.7206 & 15 & 0.7207   \\
3 & 22 & 0.7263 & 17 & 0.7241$\downarrow$ & 23 & 0.7218 & 18 & 0.7210 & 15 &  0.7207 & 12 & 0.7207 \\
4 & 21 & 0.7265 & 22 & 0.7242$\downarrow$ & 21 & 0.7220 & 14 & 0.7211 & 14 & 0.7207 & 17 & 0.7207\\
5 & 17 & 0.7272 & 14 & 0.7242$\downarrow$ & 20 & 0.7225$\downarrow$ & 17 & 0.7211 & 21 & 0.7208 & 21 & 0.7208\\
6 & 20 & 0.7275 & 23 & 0.7246$\downarrow$ & 18 & 0.7225$\downarrow$ & 15 & 0.7211 & 12 & 0.7208 & 23 & 0.7208\\
7 & 23 & 0.7282$\downarrow$ & 18 & 0.7246$\downarrow$ & 14 & 0.7226$\downarrow$ & 23 & 0.7211$\downarrow$ & 13 & 0.7209 & 9 & 0.7211\\
8 & 16 & 0.7284$\downarrow$ & 21 & 0.7247$\downarrow$ & 17 & 0.7226$\downarrow$ & 12 & 0.7212 & 23 & 0.7210$\downarrow$ & 7 & 0.7211\\
9 & 24 & 0.7286$\downarrow$ &  13 & 0.7247$\downarrow$ & 15 & 0.7226$\downarrow$ & 13 & 0.7213$\downarrow$ & 20 & 0.7210$\downarrow$ & 6 & 0.7213\\
10 & 15 & 0.7305$\downarrow$ & 24 & 0.7248$\downarrow$ & 12 & 0.7227$\downarrow$ & 20 & 0.7213$\downarrow$ & 10 & 0.7211 & 4 & 0.7215 \\
\hline
\textit{Layers Included}& \multicolumn{2}{c|}{\it{--}} & \multicolumn{2}{c|}{\it{19}} & \multicolumn{2}{c|}{\it{19;16}} & \multicolumn{2}{c|}{\it{19;16;24}} & \multicolumn{2}{c|}{\it{19;16;24;22}} & \multicolumn{2}{c|}{\it{19;16;24;22;18}}\\
\hline
\end{tabular}
\caption{\label{tab:layer_comparare} Comparison of performance between the top 10 best individual layers and additional layers on the 10-fold cross validation data, ordered by mean squared error. \textbf{Bold} indicates best in column and $\downarrow$ indicates significantly lower performing models $p < .05$ via paired t-test compared to best in column (rank 1). The best performing of the previous column is used to find the next best layer to add on (via concatenation indicated by ;). During cross-folds training N=16,694 and validation N=905.}
\end{small}
\end{center}
\end{table*}

\paragraph*{Regression: }

Our model of choice is a regularized linear regression (ridge) with input being the mean aggregate of extracted contextual embeddings. To find the regularization parameter $\alpha$, we use a 10-fold cross-validation technique searching between 10 and 1 million, increasing by powers of 10 each time, then selecting the $\alpha$ that gave the lowest mean squared error. A simple predictive model is chosen to highlight the improvements from the features themselves rather than any specific network architecture. 

\section{Dataset \& Baseline}

\paragraph*{Dataset: }
The dataset is comprised of Facebook users who opted in to share their status updates between 2009 and 2011 and completed a personality questionnaire~\cite{schwartz2014towards}. There are ~25,000 train users and 1,000 test users which are then filtered down to those who wrote at least 1,000 words across all of their status updates. The final result is a training set of 17,599 and test set of 986 users.  
\paragraph*{Baseline:}

We compare to the proposed model of ~\citet{schwartz2014towards} which leverages both open-vocab and count based lexicons. Notably, the model is trained on 1 - 3 grams, a 2000 dimensional social media LDA topic vector, Lexical Inquiry and Word Count (LIWC) lexicon, and NRC sentiment lexicon~\cite{pennebaker2001linguistic, mohammad2013nrc}. We compare our models both to the reported scores in the original publication and to a version we recreated, referred to as Open-Ridge.

\section{Evaluation}

Our recreated Open-Ridge came within .019 $r_{dis}$ of the original work, however, both the recreated and original model are outperformed by both BERT and RoBERTa base variants, as shown in table \ref{tab:baseline_base}. Interestingly ALBERT, while being 10x smaller than the other language models, performs quite well; outperforming XLNET and baseline models. We also see that all models based on the autoencoder style architecture (BERT variant) perform better than autoregressors (XLNet). This suggests that for human-level analysis the autoencoder style models are better than autoregressors, agreeing with the findings of ~\citet{v-ganesan-etal-2021-empirical}.

We also compare against possible variants of RoBERTa, which offer a computation versus performance trade-off, RoBERTa-large (24 layers) and DistilRoBERTa (6 layers) in table \ref{tab:roberta_compare}. Ultimately, RoBERTa-large performs only slightly better than the base model. While this small difference is found to not be statistically significant, due to the number of available layers of RoBERTa-large this gives more options for layer selection without a loss in performance and move forward with RoBERTa-large as our selected model.

\begin{table}[tb]
\centering
\begin{tabular}{lcc}
\hline
    \textbf{Layer Combo} &\textbf{$r_{dis}$} & \textbf{MSE}\\ \hline
    \textit{Standard}&  \\
    \hspace{14pt}  L23 & .542 & .7476 \\ 
    \hspace{14pt}  L21+22+23+24 & .546 & .7479 \\
    \textit{Optimized}&  \\
    \hspace{14pt}  L19 & .553 & .7439 \\
    \hspace{14pt}  L16+19+22+24 & .552* & .7208* \\
    \textit{Other Sizes}&  \\
    \hspace{14pt}  L16+18+19+22+24 & \textbf{.554*}& \textbf{.7206*} \\
    \hspace{14pt}  L14+16+18+19+22+24 & .553* & .7433* \\
    \hline
    \end{tabular}
    \caption{Performance of extracting embeddings using standard techniques and from the optimized layers we find to be most promising via cross-fold selection. \textbf{Bold} indicates best in column and * indicates statistical significance $p < .05$ w.r.t standard top-4 (21-24) layer extraction via paired t-test.}
\label{tab:test_results}
\end{table}


As mentioned in section \ref{sec:methods}, for investigating layer selection we only evaluate on cross-fold validation results to avoid any overfitting to the test set. First, we look at all individual layers of RoBERTa, as shown in in figure \ref{fig:roberta_layers}, and the standard errors associated with each layer's performance across the 10 cross-folds. We find that performance slowly improves as you move up the model but begins to slow down around the middle layers and peaks at layer 19.

Next, we explore the question of how many layers should be used as well as which layers to extract in order to build a user representation. For this, we apply our layer selection technique based on empirical results of the cross-folds. We show results for the top 10 best combinations per layer amount in table \ref{tab:layer_comparare}. We find 3 interesting outcomes from our experiments: (1) When using only a single layer the second-to-last is not the best and is not even in the top 5, (2) We do not see a drop in performance from using more than 4 layers, in fact, we do not see a plateau until we try 6 total layers thus suggesting that for human-level predictions large representations are ideal and (3) The layers that boost performance all come from the top half of RoBERTa-large likely due to them including more semantic information than syntactic~\cite{rogers2020primer}, which could be more informative for modeling at the human-level. 

Lastly, we compare our optimized extraction models to the standard approaches on the held-out test set; shown in table \ref{tab:test_results}. We find that our layer 19 model performs quite well but is not a statistically significant finding (p=.08) when compared against layer 23. Our 4-layer model continues to give a boost in performance and is found to be statistically significant compared to standard top-4 extraction. The 5-layer version has a small improvement in both metrics and is found to be significant(p=.02) compared to our optimized 4-layer model. For the 6-layer model we see an expected drop in performance, based on cross-fold analysis, suggesting that the additional layer has hurt the model's ability to generalize. 




\section{Conclusion}

With many tasks in NLP focused around human-level prediction, methods that can use state-of-the-art, off-the-shelf models in the best way are of interest to the community at large. In this work, we found that applying pre-trained transformer language models to depression assessment benefited from non-standard extraction techniques. Further, applying a straight forward empirical analysis of layer performance could lead to noticeable boosts in downstream applications. Ultimately, we achieved sate-of-the-art performance of $r_{dis}=.554$ and $MSE = .7206$ using a 5-layer user representation from RoBERTa-large. 

\paragraph{\textit{Ethics Statement: }} Our work is part of a growing body of interdisciplinary research that aims to improve the automatic assessment of a person's mental health. However, at this time we do not suggest our model(s) be used in practice to label mental health states. Instead, this should be viewed as a step toward a clinical tool that would be used with professional oversight. This research has been approved (deemed exempt status) by an academic institutional review board.

\bibliography{anthology,custom}
\bibliographystyle{acl_natbib}

\appendix

\end{document}